\documentclass[12pt,journal,compsoc,onecolumn]{IEEEtran}
\usepackage{amsmath,graphicx,multirow}
\usepackage{bbm,amssymb}
\usepackage{color}
\usepackage{dsfont}

\usepackage[ruled,lined,linesnumbered,boxed]{algorithm2e}
\usepackage{algorithmic}
\usepackage[noadjust]{cite}

 \def\y{{\bf y}}
 \def\x{{\mathbf x}}

 \def\w{{\mathbf w}} 
 \def\U{{\mathbf U}} 
  
 \def\A{{\mathbf \Lambda}}

 \newtheorem{proposition}{Proposition}

\title{End-to-end training of deep kernel map networks for image classification} 

\author{Mingyuan Jiu$^{\star,\dagger}$ \qquad \thanks{This work was supported by the National Natural Science Foundation of China (No.~61806180), Key Research Projects of Henan Higher Education Institutions in China (No.~19A520037) and Zhengzhou Municipal Science Technology Innovation Project (No.~2019CXZX0037), and also in part by a grant from the research agency ANR (Agence Nationale de la Recherche) of France under the MLVIS project (ANR-11-BS02-0017).}
Hichem Sahbi$^{\ddagger}$ \qquad \\
\vspace{1cm} 
{\small {$^{\star}$ School of Information Engineering, Zhengzhou University, Zhengzhou, China\\
$^{\dagger}$ Zhengzhou University Research Institute of Industrial Technology Co., Ltd., Zhengzhou, China\\
$^{\ddagger}$ CNRS LIP6 Lab, UPMC, Sorbonne University, Paris, France}}
}
\begin{document}
\maketitle
\begin{abstract}

Deep kernel map networks have shown excellent performances in various classification problems including image annotation. Their general recipe consists in aggregating several layers of singular value decompositions (SVDs) -- that map data from input spaces into high dimensional spaces -- while preserving the similarity of the underlying kernels. However, the potential of these deep map networks has not been fully explored as the original setting of these networks focuses mainly on the approximation quality of their kernels and ignores their discrimination power.  \\
\indent In this paper, we introduce a novel ``end-to-end'' design for deep kernel map learning that balances the approximation quality of kernels and their discrimination power.  Our method  proceeds in two steps; first, layerwise SVD is applied in order to build initial deep kernel map approximations and then an ``end-to-end'' supervised learning is employed to further enhance their discrimination power while maintaining their efficiency. Extensive experiments, conducted on the challenging ImageCLEF annotation benchmark, show the high efficiency and the out-performance of this two-step process with respect to different related methods.
\end{abstract}
\hspace{1cm}{\small {\bf Keywords---}   Deep kernel networks, deep map networks, supervised end-to-end learning, image annotation.}
 
\section{Introduction} \label{sec:intro}
Kernel learning has been an active research field in the last two decades with many applications ranging from support vector classification~\cite{Caputo2004,sahbi2002scale,sahbi2004kernel,lingsahbieccv2014, lingsahbiicip,joachims2009,ingo2009,sahbijmlr06,Hsu2002,weston98,benett99,icml08,lingsahbi2013,sahbiaccv2010,sahbicvpr08a} to regression~\cite{sahbijstars17,gunn98,awad2015,smolla2004}, through dimensionality reduction~\cite{ShaweTaylor2004,dutta2017high,sahbi2008particular,dimred2004,kenji2004, Ham2004,fuzzy05,hoff2007,mika1999,sahbikpca06,Lin2010,st2004,rosanna99,hoff2008,Kilian2005,sahbiicip09,liu2004,rathi2006}. More recently, an extension of kernels known as deep kernel networks (DKNs) has attracted a particular attention~\cite{Cho2009, sahbiicassp16b,Yu2009nips, jiu2015semi,Zhuang2011a, sahbiiccv17,Mairal2014, Jiutip2017} following the resurgence of neural networks~\cite{Hinton2006, Villegas2013}. These deep kernels -- defined as nonlinear and recursive combinations of standard positive semi-definite (p.s.d) kernels~\cite{Jiutip2017,sahbiicassp16b} -- are proven to be successful in describing and comparing highly nonlinear data. However, the downside of DKNs resides in their limited efficiency; indeed, the computational complexity of these networks scales linearly w.r.t. their depth and quadratically w.r.t. the size of training data, and this makes their evaluation clearly intractable for large (and even mid) scale problems. \\

\indent \textcolor{black}{According to the kernel theory (see for instance~\cite{Vapnik1998}), any p.s.d kernel admits an implicit or explicit map in a high (possibly infinite) dimensional Hilbert space.} Considering this property, an interesting alternative to kernels, is to design their associated maps explicitly. In the related work, kernel map approximation techniques include (i) the Nystr\"{o}m expansion~\cite{Williams2001} which obtains low-rank kernel maps using uniformly sampled data without replacement, (ii) the random Fourier sampling (proposed by Rahimi and Recht \cite{Rahimi2007} for gaussian kernels and extended to group-invariant kernels in~\cite{LiIonescu2010}) and (iii) the explicit kernel map design for additive homogeneous kernels~\cite{Vedaldi2012}. Other solutions rely rather on the strength of deep learning (such as~\cite{Jiu2016}) and aim at designing explicit kernel representations using deep map networks (DMNs); the latter make it possible to build maps whose inner products approach the original DKNs while being deep and highly efficient. A more recent extension, in~\cite{JiuPR2019}, further enhances the approximation quality of DMNs using unsupervised learning. Nevertheless, the discrimination power of DMNs has not been fully investigated in these works; indeed, these networks are biased towards the underlying DKNs and their design {\it ignores} the targeted classification tasks. \\

\begin{figure}[tbp]
\center
\includegraphics[width=0.65\linewidth]{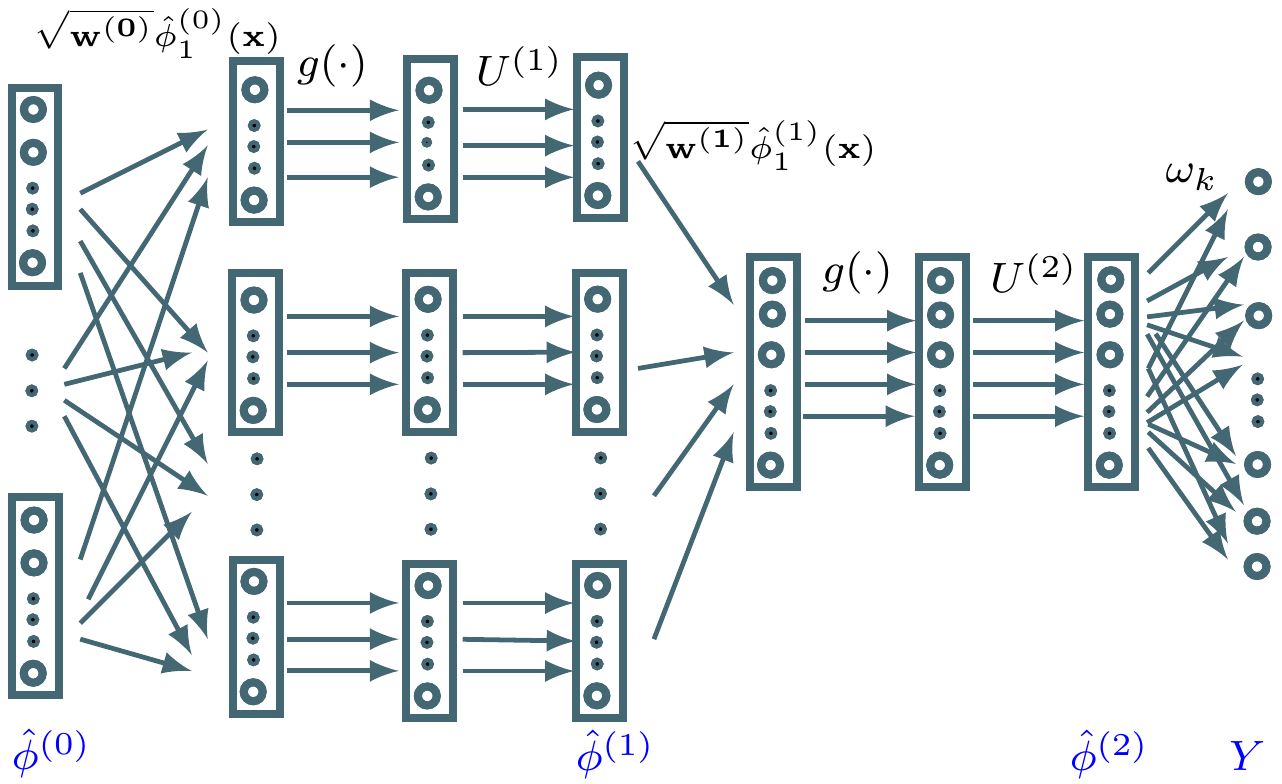}
\caption{This figure shows a three-layer deep map network (DMN); each rectangular block corresponds to the explicit map of an -- input, intermediate or output -- layer in the underlying deep kernel network (DKN). The last fully connected  layers are used for classification {\bf (better to zoom the PDF version)}. }
\label{fig:dmn}
\end{figure}

\indent Following the tremendous success of deep representation learning~\cite{Bengio2013pami,chris2017,indola2014,indola2016,mike1997,zhang2020,zaremba2014,Ishaan2017,salimans2016,Cun2015} particularly in image classification (see for instance~\cite{Krizhevsky2012, Farabet2013, SzegedyCVPR2015, HeCVPR2016, HuangCVPR2017}), we introduce in this paper an ``end-to-end'' framework that further enhances the discrimination power of the learned DMN representations. Our DMN inputs (associated to standard kernels such as polynomial kernel) are forwarded to output layers through intermediate projection and activation operations. In contrast to the aforementioned kernel approximation techniques which are mainly unsupervised, our formulation is both ``deep and supervised'' and proceeds by a greedy (layerwise) SVD decomposition followed by an ``end-to-end'' supervised learning that improves the discrimination power of the DMNs while maintaining their approximation quality w.r.t. the underlying DKNs. Note that our work is related to the method in~\cite{Mairal2014,Julien2016superckn} which approximates the maps of gaussian kernels using convolutional neural networks and the recent work of deep neural mapping  proposed by Li and Ting~\cite{LiNeural2017} which combines SVMs and kernel maps. Our method is also related to~\cite{Mccane2017deep} which proposes deep radial kernel networks to approximate gaussian kernel SVMs and also the method in \cite{Mehrkanoon18nc} that considers a deep hybrid neural-network based on random Fourier features combining neural networks and kernel machines. However, these related methods as well as those cited earlier, are either shallow or restricted to a sub-class of kernels (including gaussians) while {\it our solution in this paper is deep and targeted to a more general class of deep nonlinear kernels}. 

\section{Deep kernel map networks} \label{sec:dmn}
A deep kernel is defined as a multilayered network whose units  -- denoted as $\{\kappa_p^{(l)}\}_{l,p}$ --  correspond to (input or intermediate) kernels with $\kappa_p^{(l)}(.,.)=g(\sum_{q=1}^{n_{l-1}} \w^{(l-1)}_{p,q} \kappa_{q}^{(l-1)}(.,.))$; here $g$ is a nonlinear activation, $p$ refers to the $p$-th unit of the $l$-th layer, with $l \in \{1,\dots,L\}$, $p \in \{1,\dots,n_l\}$ and  $q \in \{1,\dots,n_{l-1}\}$. Considering p.s.d input kernels $\{\kappa_p^{(1)}\}_{p}$ (such as linear, polynomial and gaussian), provided that the weights $\{\w^{(l-1)}_{p,q}\}_{l,p,q}$ are positive and resulting from the closure of the p.s.d  w.r.t. sum and product, any intermediate kernel is at least conditionally  p.s.d for a particular class of activation functions (including hyperbolic tangent and exponential). Extra details about the setting of these weights together with the activation functions that guarantee the conditional positive semi-definiteness can be found in ~\cite{Jiutip2017}. \\

\indent \textcolor{black}{Considering the above definition, a conditionally p.s.d kernel $\kappa_p^{(l)}$ admits an explicit (either exact or approximate) map $\hat{\phi}_p^l(\cdot)$ s.t. $\kappa_{p}^{(l)}(\x,\x') \simeq \langle \hat{\phi}_p^l(\x), \hat{\phi}_p^l(\x')\rangle$; here $\hat{\phi}_p^l(\x)$ is a mapping that takes $\x$ from an input space into a high dimensional Hilbert space.} Let ${\cal S}=\{\x_i\}_{i=1}^N$ denote $N$ samples taken from our training set; assuming the kernel maps $\{\hat{\phi}_p^1(\cdot)\}_p$ of the first layer known (either exactly or tightly approximated~\cite{Sahbi2015}), we recursively define the explicit map  $\hat{\phi}_p^l(\cdot)$ of the $p$-th unit and the $l$-th layer as
\begin{equation} 
\hat{\phi}_p^{(l)}(\x)^{{\top}}= \bigg(g(\langle \hat{\phi}_p^{l,c}(\x), \hat{\phi}_p^{l,c}(\x_1) \rangle) \dots g(\langle \hat{\phi}_p^{l,c}(\x), \hat{\phi}_p^{l,c}(\x_N) \rangle)\bigg) \U_p^{(l)},
\label{equa:projection}
\end{equation}
where $^\top$ denotes the matrix transpose operator and $g(\cdot)$ stands for an activation function taken, in practice, as hyperbolic for intermediate layers and exponential for the final layer, and
\begin{equation} 
\hat{\phi}_p^{l,c}(\x) =\bigg(\sqrt{\w_{{p,1}}^{(l-1)}} \hat{\phi}_1^{(l-1)}(\x)^{{\top}} \cdots \sqrt{{\w^{(l-1)}_{p,n_{l-1}}}} \hat{\phi}_{n_{l-1}}^{(l-1)}(\x)^{{\top}}\bigg)^{{\top}}.
\label{equ:fullfeature}
\end{equation}
In Eq.~(\ref{equa:projection}), $\U_p^{(l)}={\bf V} \A^{-1\slash2}$ \textcolor{black}{is a transformation matrix} obtained by solving the following eigenproblem  
\begin{equation}
{\bf K}_p^l {\bf V}  = {\bf V} \A,
\label{equ:eigenvalueproblem}
\end{equation}
here $\mathbf{K}_p^l$ is the kernel matrix of $\kappa_p^{(l)}$ on ${\cal S}$. Fig.~\ref{fig:dmn} shows an example of a DMN architecture obtained using Eqs.~\eqref{equa:projection} --~\eqref{equ:eigenvalueproblem}. Considering the maps of these equations one may introduce the following proposition.
\begin{proposition} \label{prop:two} 
Let ${\cal S}=\{\x_i\}_{i=1}^N$ be a subset of $N$ samples and let ${\bf K}_p^{l}$ be a gram-matrix whose entries are defined on $\cal S$. Let $\U_p^{(l)}= {\bf V} \A^{-1\slash2}$ with ${\bf V}$, $\mathbf{\Lambda}$ being respectively the matrices of eigenvectors and eigenvalues obtained by solving Eq.~(\ref{equ:eigenvalueproblem}).
Considering $\|.\|_2$ as the $\ell_2$ (matrix) norm and $\hat{\bf K}_p^l$ as the gram-matrix associated to $\{\langle \hat{\phi}_p^{(l)}(\x), \hat{\phi}_p^{(l)}(\x')\rangle\}_{\x,\x' \in {\cal S}}$ with Eq.~(\ref{equa:projection}) and Eq.~(\ref{equ:fullfeature}),
\noindent  then the following property is satisfied \begin{equation}\big\|\hat{\bf K}_p^l - {\bf K}_p^l\big\|_2=0.\end{equation} 
\end{proposition}
\noindent Details of the proof are omitted and can be found online\footnote{https://www.dropbox.com/s/pdpixj73xwxjevz/suppIcip2020.pdf?dl=0}. More importantly, this proposition shows that the inner products obtained using kernel maps in Eqs.~\eqref{equa:projection} --~\eqref{equ:eigenvalueproblem} are equal to the original kernel values if data belong to $\cal S$; otherwise one may at least show that $\big\|\hat{\bf K}_p^l - {\bf K}_p^l\big\|_2 \leadsto 0$ when $N$ is sufficiently large.
\section{End-to-end Learning} \label{sec:endtoend}
\begin{algorithm}[t]
\caption{End-to-end supervised DMN learning} \label{algo:endtoend}
\BlankLine
\KwIn{Parameter setting; set the learning rate $\eta>0$. \\
Initialization: $ \{\w_{p,q}^{(l)}\}, \{\hat{\phi}_p^{l,c}(\x_i)\}_i, \{\U_{p,q}^{(l)}\}, \{\omega_k\}_k$, \\  $l \in \{1, \ldots, L\}, \  i \in \{1, \ldots, N\}, \  k \in \{1,\ldots, K\}$.} 

\Repeat{Convergence}
{
Optimize $\{\omega_k\}_k$ by LIBLINEAR toolbox\;
Compute the gradients $\frac{\partial E}{\partial \hat{\phi}_1^{L}({\bf x}_i)}$ by Eq.~\eqref{equa:gradientsvm}\;
Compute the gradients $\Delta \U_p^{(l)}$, $\Delta \hat{\phi}_p^{l,c}(\x_i)$ and $\Delta \mathbf{w}_{p,q}^{(l)}, \forall l \in \{L-1, \ldots, 1\}$\; 
Update these parameters by gradient descent:
$\U_p^{(l)} \leftarrow  \U_p^{(l)} - \eta \Delta \U_p^{(l)}$ \;
$\hat{\phi}_p^{l,c}(\x_i) \leftarrow \hat{\phi}_p^{l,c}(\x_i) - \eta \Delta \hat{\phi}_p^{l,c}(\x_i)$\;
$\mathbf{w}_{p,q}^{(l)} \leftarrow  \mathbf{w}_{p,q}^{(l)} - \eta \Delta \mathbf{w}_{p,q}^{(l)}$ \;
}
\end{algorithm}

As designed above, DMNs significantly reduce the computational complexity of DKNs especially on large scale datasets. However, with this particular unsupervised setting, DMNs fit their DKN counterparts but ignore the underlying classification tasks. Besides, kernel map design as shown in Eqs.~\eqref{equa:projection} and~\eqref{equ:fullfeature} considers only $\{\hat{\phi}_p^{l,c}(\x_i)\}_{i,l,p}$ and $\{\U_p^{(l)}\}_{l,p}$ as training parameters of DMNs while  $\{\w_{p,q}^{(l)}\}_{l,p,q}$ are fixed and taken from  DKNs. As a result, the potential of DMNs is not fully explored for supervised classification. In what follows, we propose an end-to-end supervised framework that further enhances  the discrimination power of DMNs.\\

Let $\mathcal{T} = \{({\bf x}_i,\y_i^k)\}_i$ be a training set whose samples belong to  $K$ classes; here  ${\bf x}_i$ is a training data and ${\y}_i^k$ its class membership, with  ${\y}_i^k=+1$ iff ${\bf x}_i$ belongs to class $k$, otherwise ${\y}_i^k=-1$. \textcolor{black}{For classification, a fully connected layer with $K$ units is stacked on top of the DMN whose parameters $\{\mathbf{\omega}_k\}_{k=1}^K$} correspond to the normals of $K$ hyperplane classifiers. The decision function of each classifier $f_k$ is given by
\begin{equation}
f_k(\mathbf{x}_i) = \omega_k^\top \hat{\phi}_1^{L}(\mathbf{x}_i).
\end{equation}
\noindent With this decision function, a concept $k$ is declared as present in $\x_i$ iff the score $f_k(\mathbf{x}_i)$ is positive. In order to learn the DMN, we minimize a squared hinge loss criterion rather than the widely used logistic loss, due to the fact that the former has Lipschitz continuous gradients and shows good discrimination ability for classification~\cite{TangICML2013}. Therefore, the loss $E$ to minimize is written as  
\begin{equation}\label{equa:hingeloss}
\min_{\hat{\phi}_p^{l,c}, \U_p^{(l)}, \w_{p,q}^{(l)}, \mathbf{\omega}_k} \sum_{k=1}^K \frac{1}{2} ||\omega_k||_2^2 + C_{k} \sum_{i} \max \big(0, 1-\y_i^k f_k(\mathbf{x}_i)\big)^2,
\end{equation} 
\noindent where the first term is an $\ell_2$ penalization, the second one is an empirical loss on training data and $C_k$ controls the influence of these two terms.\\

\textcolor{black}{In order to minimize Eq.~\eqref{equa:hingeloss}, we adopt an alternating optimization strategy that makes training tractable: first, we optimize the classifier weights by LIBLINEAR~\cite{LinLinsvm2008} while fixing the parameters of DMN and then we update the latter while fixing the classifier weights. In the second step, when classifier weights are fixed, the gradient of $E$ w.r.t. the output of DMN (i.e.~$\hat{\phi}_1^{L}(.)$) is given as
\begin{equation} \label{equa:gradientsvm}
 \frac{\partial E}{\partial \hat{\phi}_1^{L}({\bf x}_i)} = -2 \displaystyle  \sum_{k=1}^K C_k \y_i^k \omega_k \max \big(0, 1-\y_i^k f_k(\mathbf{x}_i)\big).
\end{equation}
\noindent Then, we employ the chain rule~\cite{LeCun98} and we back-propagate the above gradient to the preceding layers in DMN to obtain the gradients of $E$ w.r.t.~$\{\w_{p,q}^{(l)}\}_{l,p,q}$, $\{\hat{\phi}_p^{l,c}(\x_i)\}_{i,l,p}$ and $\{\U_p^{(l)}\}_{l,p}$. Finally, we update the DMN parameters using gradient descent. The learning procedure is repeated till convergence or when the maximum number of iterations is reached (see more details in Algorithm~\ref{algo:endtoend})}. 
\begin{figure*}[ht]
\begin{center}
\includegraphics[angle=0,width=0.9\linewidth]{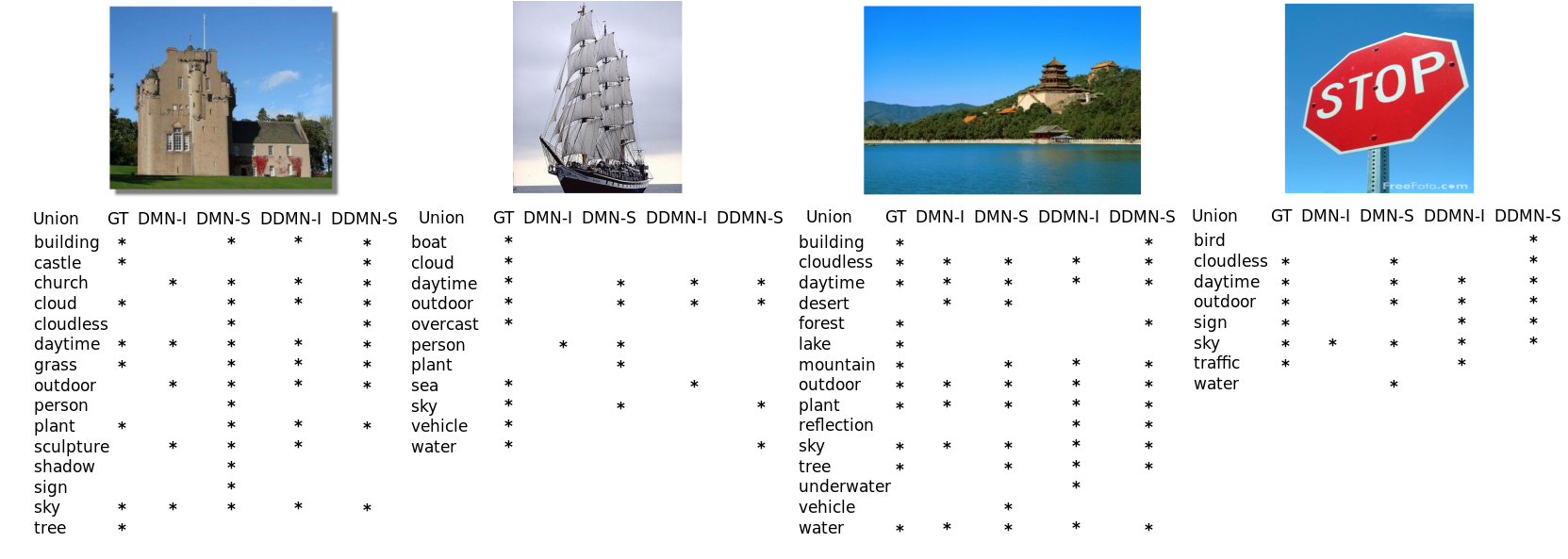}
\end{center}
\caption{This figure shows examples of annotation results using the original and the end-to-end DMNs (resp. denoted as ``DMN-I'', ``DMN-S'' for  handcrafted features and ``DDMN-I'',  ``DDMN-S'' for deep features, ``I'' for initialization and ``S'' for supervised end-to-end learning.). ``GT'' stands for ground truth annotation while the stars mean the presence of a concept in a test image.} \label{fig:annotationexamples}
\end{figure*}

\section{Experiments} \label{sec:experiments}
In this section, we evaluate the performance of the proposed algorithm on the challenging and widely used ImageCLEF annotation benchmark. The goal of image annotation (also known as  multi-label classification) is to predict the presence of semantic concepts into images; note that concepts in this task are not exclusive, so each image may be annotated by one or multiple concepts and this makes image  annotation a highly challenging task (see also \cite{vo2012transductive,sahbi2003coarse,sahbi2000coarse,eve2005, ross2015,sahbipr2012,javad2017,He2017,sahbiigarss12b,ozu2016,sahbiigarss12,sahbispie2004,icip2001,sahbiclef08,sahbicassp11,temporalpyramid,sahbiclef13,Zheng2018,Bahrololoum2017,Bhagat2018,Pellegrin2017,Andrade2018,Firmino2019,Bouzaieni2017,Heidy2019,Murthy2016,Zang2017,Xiao2019,Xiao2017,Yanchun2019,Yulei2018,Cheng2018}).\\

\indent ImageCLEF dataset~\cite{Villegas2013} contains more than 250k (training, dev and test) images belonging to 95 different concepts. We only use the dev set of 1000 images in our experiments, as the ground truth is released only for this subset. The dev set is split into two subsets: the first one is used for training, and the other for testing. The discrimination power for annotation task are evaluated using F-measures (harmonic means of recalls and precisions) both at the image and the concept levels (denoted as MF-S and MF-C respectively) as well as mean average precision (mAP). Higher values of these measures imply better performances.\\

Firstly, we study the performance of ten handcrafted visual features (provided by the ImagCLEF challenge organizers). Four input kernels (linear, polynomial, RBF and histogram intersection) are considered for each feature, so we have 40 input kernels in total. Then we learn a three-layer DKN with a hidden layer of 80 units (i.e.~twice the size of the input as set in~\cite{Jiutip2017}), next we build an initial DMN by using the method in~\cite{Jiu2016}. Finally, we update this DMN using the proposed end-to-end algorithm. The classification weights in the final layer are randomly initialized, the trade-off parameter $C_k$ is initially set using 3-fold cross-validation on the training subset and the learning rate is empirically set to $10^{-6}$ to guarantee  convergence. In these experiments, we observe that 500 iterations are sufficient in order to converge to a stable solution. Tab.~\ref{tab:results} shows the comparison between the DKN, the initial DMN and the enhanced (end-to-end) DMNs on the handcrafted features. 

Secondly, we conduct another set of experiments by taking into account  the deep features from the pre-trained VGG model on the ImageNet database (``imagenet-vgg-m-1024'')~\cite{Chatfield14}, containing five convolutional layers and three fully-connected layers. We use the outputs of the second fully-connected layer in order to describe images. Similarly, we consider four input kernels on top of the deep features, and we repeat the experiments as described above. The comparative results are again shown in Tab.~\ref{tab:results}; the latter shows performances for different settings including the original DKN, its two initial DMNs (trained using SVD, with and without unsupervised learning), and the two underlying ``end-to-end'' DMN variants whose weights are initially taken from these two initial DMNs respectively. \\

\begin{table}[tb]
	\centering
\resizebox{0.55\textwidth}{!}{	
	\begin{tabular}{c|c|ccc}
	\hline
	Features & Method & MF-S & MF-C & MAP \\
    \hline
    & GMKL(\cite{Varma2009})  & 41.3 & 24.3 & 49.1 \\
    & 2LMKL(\cite{Zhuang2011a}) & 45.0 & 25.8 & 54.0 \\
    & LDMKL (\cite{Jiutip2017}) & 47.8 & 30.0 & 58.6 \\
    \cline{2-5}
    & DKN (\cite{Jiutip2017}) & 46.2 & 30.0 & 55.7  \\
    Handcrafted & Ini. DMN (\cite{Jiu2016}) & 47.7 & 29.4 & 53.2  \\
    feat. & Sup. DMN (Proposed) & 49.6 & 31.9 & 58.5  \\
    & Uns. DMN (\cite{JiuPR2019}) & 48.0 & 29.8 & 53.3 \\
    & Ft. DMN (Proposed) & 50.2 & 32.2 & 59.2 \\
	\hline
    & DKN (\cite{Jiutip2017}) & 56.3 & 38.9 & 66.6 \\
    +Deep  & Ini. DMN (\cite{Jiu2016}) & 56.7 & 39.7 & 66.4 \\
    feat. & Sup. DMN (Proposed)  & 56.8 & 40.4 & 67.2 \\ 
    & Uns. DMN (\cite{JiuPR2019}) & 56.4 & 39.3 & 66.5 \\  
    & Ft. DMN (Proposed) & 56.2 & 40.5 & 67.4 \\    
	\hline	
	\end{tabular}
	}
	 \vspace{0.5cm} 
        \caption{  \textcolor{black}{Comparison of annotation performances (in \%) of different methods, using handcrafted and deep features. In this table, ``Ini.'' stands for initialization, ``Sup.'' for supervised, ``Uns.'' for unsupervised and ``Ft.'' for  fine-tuned. } \label{tab:results}}
\end{table}

\begin{table}[tb]
	\centering
\resizebox{0.59\textwidth}{!}{	
	\begin{tabular}{c|c|ccc}
	\hline
	Framework & \textcolor{black}{Sample size $|{\cal T}|$} & Average runtime ($\textrm{in seconds}$) \\
    \hline
    \multirow{4}{*}{DKN} & 500  & 0.305 \\
                         & 1000 & 0.826 \\
                         & 2000 & 2.822 \\
                         & 5000 & 16.188 \\
    \hline
    \multirow{4}{*}{DMN} & 500  & 0.569 \\
                         & 1000 & 0.566 \\
                         & 2000 & 0.595 \\
                         & 5000 & 0.594 \\
    \hline
	\end{tabular}
	}
	\vspace{0.5cm}
         \caption{ This table shows the average runtime in order to classify any given sample using DKNs vs DMNs. When classifying a sample with DKNs, all the kernel values between that sample and $\cal T$ should be evaluated prior to classification using the dual SVM form; in contrast, DMNs rely on {\it efficient} explicit kernel map (+ primal SVM) evaluations. These performances were obtained on a workstation with four Xeon CPUs of 3.2GHz. \textcolor{black}{In all these experiments $|{\cal S}|=N=1000$.}} \label{tab:time-results}
\end{table}

From Tab.~\ref{tab:results}, we observe that our proposed ``end-to-end'' learning framework is able to further boost the discrimination power of DMNs compared to initial and unsupervised DMNs as well as the original DKNs on both handcrafted and deep features. Recall that ``Ini/Uns'' DMNs shown in Tab.~\ref{tab:results} are designed only to fit the underlying DKNs without taking into account any label information. In contrast, ``Sup/Ft'' DMNs make it possible to retrain their parameters while also maximizing classification performances. \textcolor{black}{Extra comparisons against other kernel-based methods are also shown in Tab.~\ref{tab:results} and they validate the effectiveness of the proposed algorithm. Fig.~\ref{fig:annotationexamples} shows several annotation results in the test set for the initial DMNs and the fine-tuned ones using supervised end-to-end learning on handcrafted and deep features. } \\ 

\indent Tab.~\ref{tab:time-results} shows a comparison of the average runtime in order to process (classify) any given sample using DKNs vs. DMNs. When using DKNs, this process requires evaluating and propagating $M$ kernel values through $L$ layers between a given  sample and all training data in $\cal T$ with a complexity  $O(M^2  L  |{\cal T}|)$; here $M=\max_{l} n_l$ (i.e., upper bound on the width of DKN). When using instead DMNs, the complexity of evaluating the kernel maps in Eqs.~(\ref{equa:projection}) and (\ref{equ:fullfeature}) is independent from $\cal T$ and equal to $O(MLN^2)$; \textcolor{black}{see also Tab.~\ref{tab:time-results}}. Hence, when $ |{\cal T}| \leq N$ and provided that $M < N$, DKNs are more efficient while larger values of $|{\cal T}|$ make DMNs more and more efficient with respect to DKNs.
\section{Conclusion}  \label{sec:concl}
We introduced in this paper an ``end-to-end'' design of deep map networks that effectively approximate the underlying deep kernel networks while being highly efficient. The strength of our method resides in its ability to fit not only the original DKNs but also the targeted classification task. Our method proceeds in two steps: first, an SVD step is achieved in order to build the initial DMN architecture, followed by an ``end-to-end'' supervised training step that further enhances the discrimination power of our DMNs and their parameters. Experiments conducted on the challenging ImageCLEF benchmark show a clear and a consistent gain of our ``end-to-end'' DMN design compared to other different settings.

\end{document}